\relax
\documentclass[letterpaper]{article} 
\usepackage{jmlr2e}
\usepackage{times,amsmath,amssymb}  
\usepackage{helvet} 
\usepackage{courier}  
\usepackage{url}  
\usepackage{graphicx,subfig} 
\usepackage{natbib}  
\usepackage[switch]{lineno}
\usepackage{caption} 
\usepackage{amsmath,amsfonts,amssymb}

\title{Calibrate and Prune: Improving Reliability of Lottery Tickets Through Prediction Calibration}
\author{\name Bindya Venkatesh \email bvenka15@asu.edu \\
	\addr Arizona State University\\
	Tempe, AZ, USA
	\AND
	\name Jayaraman J. Thiagarajan \email jjayaram@llnl.gov \\
	\addr Lawrence Livermore National Labs\\
	Livermore, CA,USA
	\AND
	\name Kowshik Thopalli \email kthopall@asu.edu \\
	\addr Arizona State University\\
	Tempe, AZ, USA
	\AND
	\name Prasanna Sattigeri \email psattig@us.ibm.com\\
	\addr IBM Research AI\\
	NY, USA
}
\begin{document}

\maketitle
\begin{abstract}

The hypothesis that sub-network initializations (lottery) exist within the initializations of over-parameterized networks, which when trained in isolation produce highly generalizable models, has led to crucial insights into network initialization and has enabled efficient inferencing. Supervised models with uncalibrated confidences tend to be overconfident even when making wrong prediction. In this paper, 
for the first time, we study how explicit confidence calibration in the over-parameterized network impacts the quality of the resulting lottery tickets. More specifically, we incorporate a suite of calibration strategies, ranging from mixup regularization, variance-weighted confidence calibration to the newly proposed likelihood-based calibration and normalized bin assignment strategies. Furthermore, we explore different combinations of architectures and datasets, and make a number of key findings about the role of confidence calibration. Our empirical studies reveal that including calibration mechanisms consistently lead to more effective lottery tickets, in terms of accuracy as well as empirical calibration metrics, even when retrained using data with challenging distribution shifts with respect to the source dataset.


\end{abstract}

\section{Introduction}
With an over-parameterized neural network, pruning or compressing its layers, while not compromising performance, can significantly improve the computational efficiency of the inference step~\cite{dettmers2019sparse}. However, until recently, training such sparse networks directly from scratch has been challenging, and most often they have been found to be inferior to their dense counterparts. Frankle and Carbin~\cite{frankle2018lottery}, in their work on lottery ticket hypothesis (LTH), showed that one can find sparse sub-networks embedded in over-parameterized networks, which when trained using the same initialization as the original model can achieve similar or sometimes even better performance. Surprisingly, even aggressively pruned networks ($>95\%$ weights pruned) were showed to be comparable to the original network, as long as they were initialized appropriately. Such a well-performing sub-network is often referred as a \textit{winning lottery ticket} or simply a \textit{winning ticket}. 

Following this pivotal work, several studies have been carried out to understand the role of initialization, the effect of the pruning criterion used and the importance of retraining the sub-networks~\cite{zhou2019deconstructing, evci2019rigging, oneticket, desai2019evaluating, gohil2019one, ramanujan2019s} for the success of lottery tickets. In~\cite{desai2019evaluating}, Desai \textit{et al.} evaluated winning tickets under data distribution shifts, and found that the tickets demonstrated strong generalization capabilities. Similarly, in~\cite{oneticket}, the authors reported that the winning tickets generalized reasonably across changes in the training configuration. 


In this paper, the focus on the fundamental problem of winning ticket selection from an over-parameterized network and the role confidence calibration plays in it. A common pitfall with supervised models in practice is that, despite achieving high accuracy on the validation data, tend to be over-confident even while making wrong predictions, and this can lead to unexpected model behavior on unseen test data. In such cases, prediction calibration strategies are used improve the reliability of models by penalizing over-confident or under-confident predictions~\cite{berthelot2019mixmatch,berthelot2019remixmatch}. Broadly, calibration is the process of adjusting predictions to improve the error distribution of a predictive model. For the first time, we propose to study the impact of confidence calibration on the quality of the resulting lottery tickets. To this end, we explore a suite of calibration strategies, and evaluate the performance of lottery tickets, in terms of accuracy and calibration metrics, on several dataset/model combinations. In addition to studying popular calibration mechanisms from the literature, we also introduce two novel strategies namely likelihood weighted confidence calibration with stochastic inferencing, and a normalized bin assignment strategies. Finally, we investigate the generalization performance of those tickets when retrained using data characterized by real-world distribution shifts, and find that confidence calibration provides significant performance gains over the standard LTH.



\section{Lottery Ticket Hypothesis}
Formally, the process of lottery ticket training in~\cite{frankle2018lottery} can be described as follows: (i) train an over-parameterized model with initial parameters $\theta_i$ to infer final parameters $\theta_f$; (ii) prune the model by applying a mask $z \in \{0,1\}^{|\theta_f|}$ identified using a masking criterion, e.g. LTH uses weight magnitudes; (iii) Reinitialize the sparse sub-network by resetting the non-zero weights to its original initial values, i.e., $z \odot \theta_i$ and retrain. These steps are repeated until a desired level of pruning is achieved.  

\noindent \textbf{Why Does LTH Work?} The work by Zhou et. al. \cite{zhou2019deconstructing} sheds light into reasons for the success of LTH training. The authors generalized the iterative magnitude pruning in~\cite{frankle2018lottery}, and proposed several other choices for the pruning criterion and the initialization strategy.  Most importantly, they reported that retaining the signs from the original initialization is the most crucial, and also  argued that zeroing out certain weights is a form of training and hence accelerates convergence. However, these variants still require training the over-parameterized model and this does not save training computations. Consequently, in \cite{wang2019picking}, Wang \textit{et al.} computed the gradient flows of a network, and performed pruning prior to training, such that the gradient flows are preserved. Note, alternate pruning approaches exist in the literature -- in \cite{molchanov2017variational}, the authors adopted variational dropout for sparsifying networks. Lee \textit{et al.}~\cite{lee2018adaptive} improved upon this by using a sparsity inducing Beta-Bernoulli prior. 

\noindent \textbf{Is Retraining Required?} Another key finding from LTH studies is that randomly initialized, over-parameterized networks contain sub-networks that lead to good performance without updating its weights~\cite{ramanujan2019s}. Similar results were reported with Weight Agnostic Networks \cite{gaier2019weight}. These works disentangle weight values from the network structure, and show that structure alone can encode sufficient discriminatory information. Another intriguing observation from~\cite{ramanujan2019s} is that certain distributions such as \textit{Kaiming Normal} and \textit{Scaled Kaiming Normal} are considerably superior to other choices. 

\noindent \textbf{Transfer Learning using LTH:} Pruning and transfer learning have been studied before \cite{molchanov2016pruning, zhu2017prune}, however there are only a handful of works so far that have explored the connection between transfer learning and LTH. For example, in~\cite{gohil2019one} the authors  investigate the transfer of initializations instead of transferring learned representations. In particular, it was found that winning tickets from large datasets transferred well to small datasets, when the datasets were assumed to be drawn from similar distributions. This empirical result hints at the potential existence of a distribution of tickets that can generalize across datasets. In this spirit, Mehta~\cite{mehta2019sparse} introduced the \textit{Ticket Transfer Hypothesis} --  there exists a sparse sub-network ($z \odot \theta^s_f$) of a model trained on the source data, which when fine-tuned to the target data will perform comparably to a model that is obtained by fine-tuning the dense model $\theta^s_f$ directly. 


\section{Improving Winning Tickets using Prediction Calibration}
In this paper, we use the term calibration to refer to any strategy that is utilized to adjust the model predictions to match any prior on the model's behavior, e.g., error distribution. Formally, we consider a $K$-way classification problem, where $\mathrm{x} \in \mathcal{X}$ and $\mathrm{y} \in \mathcal{Y}=\{1, 2, \ldots,K \}$ denote the input data and its corresponding label respectively. We assume that the observed samples are drawn from the unknown joint distribution $p(\mathrm{x},\mathrm{y})$. The task of classifying any sample $\mathrm{x}_n$ amounts to predicting the tuple $(\hat{\mathrm{y}}_n,\hat{\mathrm{p}}_n)$, where  $\hat{\mathrm{y}}_n$ represents the predicted label and $\hat{\mathrm{p}}_n$ is the likelihood of the prediction. In other words, $\hat{\mathrm{p}}_n$ is a sample from the unknown likelihood $p(\mathrm{y}_n|\mathrm{x}_n)$, which represents the associated uncertainties in the prediction, and the label $\hat{\mathrm{y}}_n$ is derived based on $\hat{\mathrm{p}}_n$. While approximating these likelihoods has been the focus of deep uncertainty quantification techniques~\cite{gal2016uncertainty}, prediction calibration has been adopted to improve model reliability. 


In this paper, we study the impact of prediction calibration during model training on the inferred tickets~\cite{frankle2018lottery} and their generalization. It is well known that supervised models with uncalibrated confidences tend to be overconfident even while making wrong predictions~\cite{guo2017calibration}. This observation is highly relevant to LTH methods, where the most popular strategy used for selecting winning tickets is to rank the network weights based on their magnitudes. We hypothesize that, while neurons with the largest magnitude are the most useful for sub-network selection, they also present the highest risk for causing over-confidences in model predictions. Consequently, including confidence calibration as an explicit training objective will temper the influence of neurons that can eventually lead to miscalibration, as they continue to be updated in the gradient descent process. For the first time, we show that pruned tickets obtained via confidence calibration, though retrained using the same initialization as the standard LTH, leads to improved performance. While calibration is specific to a task, i.e., the calibration is not guaranteed to be preserved under transfer learning to a new task, in this paper, we show that our tickets can effectively generalize under challenging distribution shifts, for the same task. More specifically, we consider the following calibration methods in our study:
\begin{itemize}
    \item \textit{No Calibration}: This is the baseline approach where we utilize only the standard cross-entropy loss for training the model. We refer to this as \textit{Basic}.
    
    \item \textit{Mixup}: Mixup is a popular augmentation strategy~\cite{zhang2017mixup} that generates additional synthetic training samples by convexly combining random pairs of images and their corresponding labels, in order to temper overconfidence in predictions. Recently, in~\cite{thulasidasan2019mixup}, it was found that mixup regularization led to improved calibration. Formally, mixup training is designed based on Vicinal Risk Minimization, wherein the model is trained not only on the training data, but also using samples in the vicinity of each training sample. The vicinal points are generated as follows: 
\begin{equation}
    \mathrm{x} = \lambda \mathrm{x}_i + (1-\lambda) \mathrm{x}_j; \quad \mathrm{y} = \lambda \mathrm{y}_i + (1-\lambda) \mathrm{y}_j,
    \label{eqn:mixup}
\end{equation}where $\mathrm{x}_i$ and $\mathrm{x}_j$ are two randomly chosen samples with their associated labels $\mathrm{y}_i$ and $\mathrm{y}_j$. The parameter $\lambda$, drawn from a symmetric Beta distribution sets the mixing ratio.

\item \textit{Variance Weighted Confidence Calibration (VWCC)}: This approach uses stochastic inferences to calibrate the confidence of deep networks. More specifically, we utilize the loss function in~\cite{seo2019learning}, which augments a confidence-calibration term to the standard cross-entropy loss and the two terms are weighted using the variance measured via  multiple stochastic inferences. Mathematically, this can be written as:
\begin{align}
    &\mathcal{L}_{vwcc} = \sum_{i=1}^N (1 - \alpha_i)\mathcal{L}_{ce}^i + \mathcal{L}_{U}^i \\
    \nonumber &=\sum_{i=1}^N -(1 - \alpha_i)\log(p(\hat{\mathrm{y}}_i|\mathrm{x}_i)) \\ &\phantom{aaaaaaaaaaaaa}+ \alpha_i D_{KL}(\mathcal{U}(\mathrm{y})||p(\hat{\mathrm{y}}_i|\mathrm{x}_i)).
        \label{vwcc}
    \end{align}Here $\mathcal{L}_{ce}^i$ denotes the standard cross-entropy loss for sample $\mathrm{x}_i$, and the predictions $p(\hat{\mathrm{y}}_i|\mathrm{x}_i)$ are inferred using $T$ stochastic inferences for each sample $\mathrm{x}_i$, while the variance in the predictions is used to balance the loss terms. More specifically, we perform $T$ forward passes with dropout in the network and promote the softmax probabilities to be closer to an uniform distribution, when the variance is large. The normalized variance $\alpha_i$ is given by the mean of the Bhattacharyya coefficients between each of the $T$ predictions and the mean prediction.
    
    \item \textit{Likelihood Weighted Confidence Calibration with Stochastic Inferences (LWCC-SI)}: 
We propose a new calibration strategy that utilizes the estimated likelihoods, in lieu of the variance weighting, to define the confidence calibration objective. More specifically, similar to \textit{VWCC}, we apply dropout and obtain $T$ different predictions for each sample. In particular,
\begin{align}
    \nonumber &\mathcal{L}_{lwcc} = \sum_{i=1}^N \mathcal{L}_{ce}^i + \lambda \beta_i D_{KL}(\mathcal{U}(\mathrm{y})||p(\hat{\mathrm{y}}_i|\mathrm{x}_i)), 
    \\
    &\text{where }\beta_i = \bigg(1 - \max(\hat{\mathrm{y}}_i)\bigg)^{\mathbb{I}(\mathrm{y}_i = \hat{\mathrm{y}}_i)}.
        \label{lwcc-si}
\end{align}The indicator function $\mathbb{I}(\mathrm{y}_i = \hat{\mathrm{y}}_i)$ ensures that the weight $\beta_i$ is at the maximum value of $1$ when the prediction is wrong, i.e.,  enforces the softmax probabilities towards a high-entropy uniform distribution. On the other hand, when the prediction is correct, the term penalizes cases when the likelihood is low. The loss function in equation~\eqref{lwcc-si} is computed using the average prediction $p(\hat{\mathrm{y}}_i|\mathrm{x}_i)$ across the $T$ realizations.

\item \textit{Marginal Distribution Alignment (MDA)}: When a classifier model is biased and assigns non-trivial probabilities towards a single class for all samples, the resulting predictions are often unreliable. In such scenarios, we can adopt a calibration strategy wherein we discourage assignment of all samples to a single class.
\begin{equation}
    \mathcal{L}_{mda} = \sum_{i=1}^N \mathcal{L}_{ce}^i + \gamma_d \sum_{k=1}^K p_k \log\left(\frac{p_k}{\bar{h}_k}\right)
    \label{eqn:mda}
\end{equation}where $p_k$ is the prior probability distribution for class $k$ and $\bar{h}_k$ denotes the mean softmax probability for class $k$ across all samples in the dataset. Similar to~\cite{arazo2019pseudo}, we assume a uniform prior distribution, and approximate $\bar{h}_k$ using mini-batches. 



\item \textit{Normalized Bin Assignment (NBA)}: A popular metric used for evaluating calibration of classifier models is the \textit{empirical calibration error} (ECE) (definition can be found in Section 4). This metric measures the discrepancy between the average confidences and the accuracies of a model. In practice, we first bin the maximum softmax probabilities (a.k.a confidence) for each of the samples and then measure bin-wise discrepancy scores. Finally, we compute a weighted average of the scores, where the weights correspond to ratio of samples in each bin. Intuitively, assigning all samples to a high-confidence bin can lead to overconfidence compared to the accuracy of the model, while assigning all samples to a low-confidence bin will produce a under-confident model even when the accuracy is reasonable. To discourage either of these cases, we propose the following regularization:
\begin{align}
    \mathcal{L}_{nba} = \sum_{i=1}^N \mathcal{L}_{ce}^i + \gamma_n \sum_{b=1}^{B} w_b \left|\frac{N_b}{N} - \frac{1}{B} \right|, 
        \label{nba}
    \end{align}where $B$ is the total number of bins considered, $N_b$ denotes the number of samples in bin $b$ and $w_b$ is the bin-level weighting. Since the operation of counting the number of samples in each bin is not differentiable, we use a soft histogram function, and we assign larger weights to lower/higher confidence bins to avoid underconfidence/overconfidence.

\end{itemize}



\section{Empirical Studies}
We perform empirical studies with different dataset/model architecture combinations to understand the impact of prediction calibration on the winning tickets. A key design choice to be made while implementing LTH is whether to prune a fixed ratio of parameters in each layer, often referred to as local pruning, as opposed to pruning a fixed ratio of all parameters of the network, i.e. global pruning. We follow the standard experiment setup used in previous works, for each of the datasets. The other crucial component in LTH is the initialization scheme used for the weights in the pruned sub-networks. More specifically, we investigated two popular strategies namely rewinding weights to the initializations of the over-parameterized network and randomly re-initializing the tickets in every iteration. In all our experiments, we found the former strategy to produce better performance and hence we report the results for only that case. Furthermore, following the recommendation in~\cite{oneticket,frankle2018lottery}, we used late-resetting of one epoch, i.e., using the weight states after training the model for one epoch to initialize the pruned tickets in lieu of the original random initialization, for all the experiments.

\begin{figure*}[t]
	\centering
	\subfloat[S][MNIST]{\includegraphics[width=0.8\linewidth,keepaspectratio]{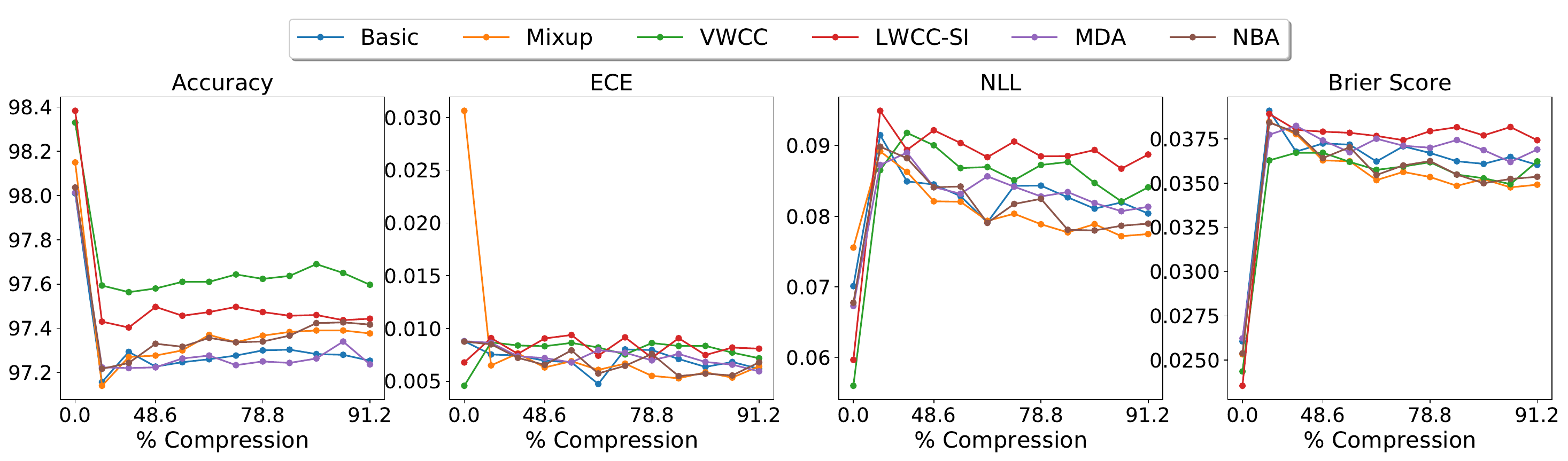} \label{fig:minist}} \vfill
	\subfloat[S][Fashion-MNIST]{\includegraphics[width=0.8\linewidth,keepaspectratio]{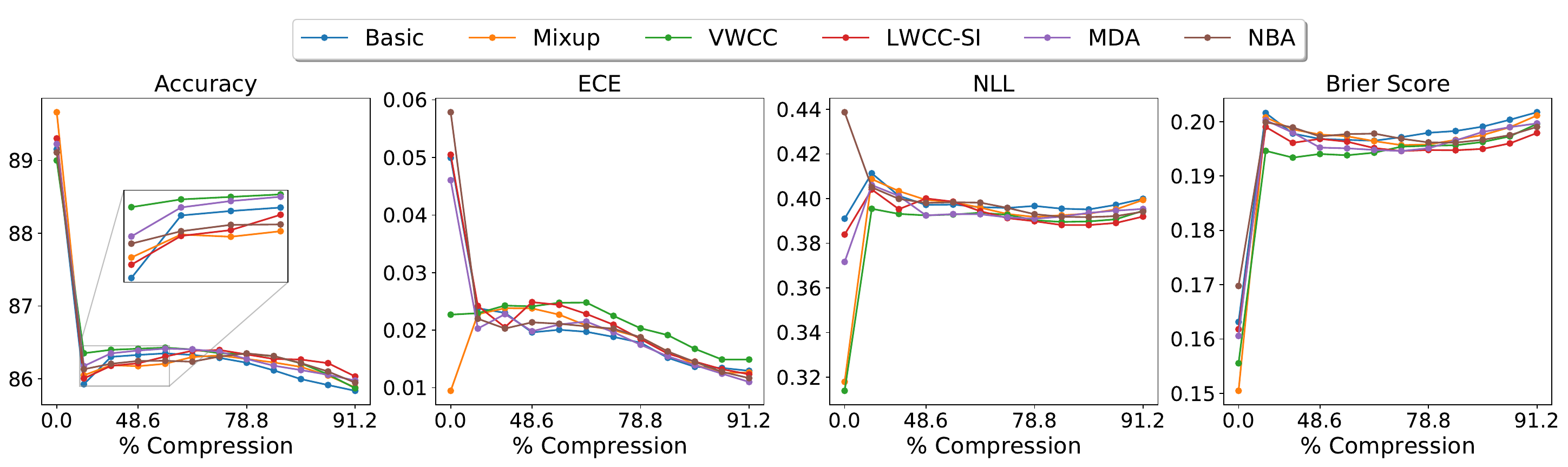}\label{fig:fm}} \vfill
	\subfloat[S][CIFAR-10]{\includegraphics[width=0.8\linewidth,keepaspectratio]{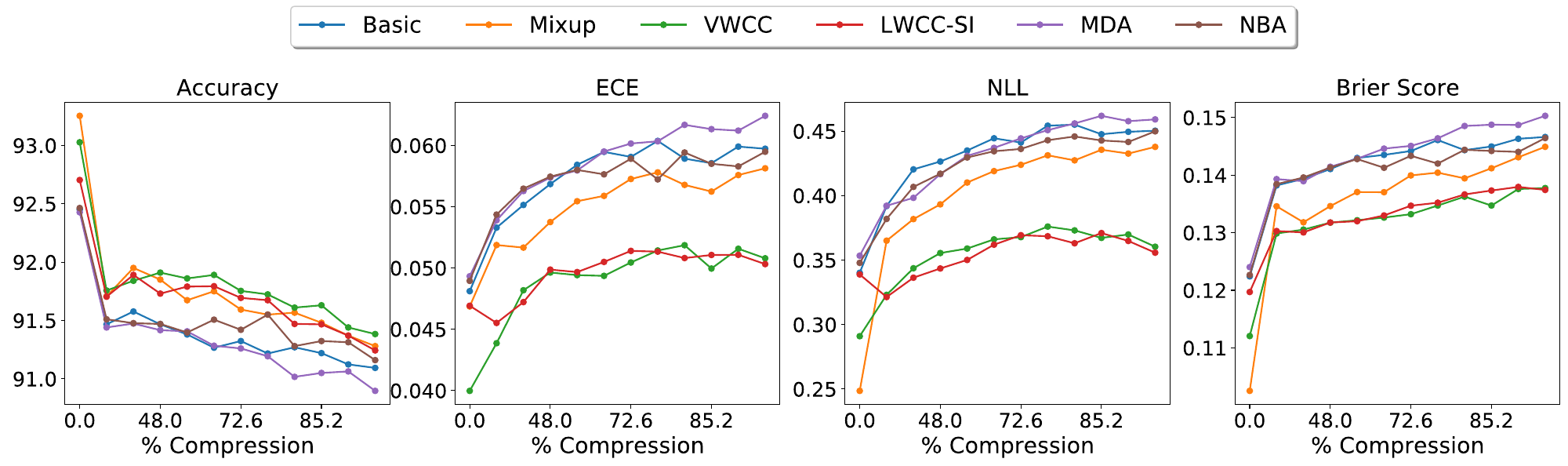} \label{fig:cifar}}
	\caption{Generalization and calibration performance of winning tickets obtained with and without an explicit calibration objective during training, for different dataset and architecture combinations - (a) A LeNet-300-100 model trained on Fashion-MNIST data; (b) A LeNet-300-100 model training on MNIST digits; (c) A ResNet-18 model trained for CIFAR-10 classification. }
	\label{fig:perf}
\end{figure*}

Though standard classification metrics such as accuracy are routinely used to evaluate the performance of lottery tickets, their reliability is not usually quantified. In a well-calibrated classifier, we expect the predictive scores to match actual likelihood of correctness~\cite{quinonero2005evaluating, guo2017calibration, degroot1983comparison}.. We use three popular calibration metrics for this evaluation, namely (i) empirical calibration error (ECE), (ii) negative log likelihood (NLL) and (iii) Brier score. We present comparisons for winning tickets obtained using different prediction calibration strategies (discussed in Section 3) while training the over-parameterized model and we report averages obtained using three different trials (random seeds). The hyperparameters used for the different calibration strategies in each of the experiments are listed in the appendix.


\noindent \textbf{Metrics.} We now formally define the calibration metrics used in our evaluation:

\noindent \textit{Empirical Calibration Error:} This is the most widely used metric to evaluate the predictions. Since ECE takes only prediction confidence into account and not the complete prediction probability, it is often considered as an insufficient metric~\cite{guo2017calibration}. Consequently, variants of this metric have been  considered~\cite{nixon2019measuring}. 
In our setup, we adopt the following strategy: we bin the maximum softmax probability (confidence) from each of the samples into $B$ bins and compute calibration error as the discrepancy between the average confidence and average accuracy in each of these bins:
\begin{equation}
\label{eq:ece}
\mathrm{ECE}=\sum_{b=1}^{B} \frac{N_{b}}{N}\left|\operatorname{acc}\left(\mathcal{B}_{b}\right)-\operatorname{conf}\left(\mathcal{B}_{b}\right)\right|,
\end{equation} where $N_b$ represents the number of predictions falling in bin $b$ and $\operatorname{acc}(\mathcal{B}_{b})$ is the accuracy and $\operatorname{conf}(\mathcal{B}_{b})$ the average confidence of the samples in bin $b$.

\begin{figure*}[t]
	\centering
	\includegraphics[width=0.8\linewidth]{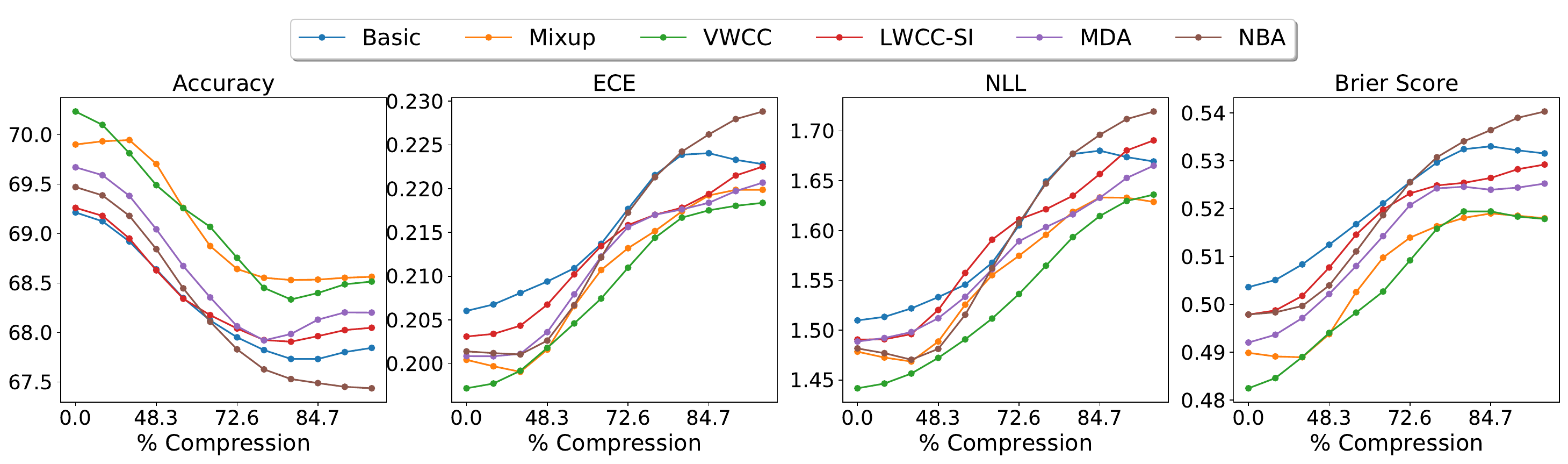}
	\caption{Ticket transfer performance on a target dataset (CIFAR-10a) from the same distribution as the source data (CIFAR-10b). Both models were implemented using the ResNet-18 architecture and we show the performance for test data in the target.}
	\label{fig:cifar10ab}
\end{figure*}

\noindent \textit{Negative Log Likelihood:} Given the prediction likelihoods, the negative log likelihood metric can be used to obtain a notion of calibration as showed in~\cite{guo2017calibration, gneiting2007strictly}. For a set of predictions on given $N$ samples, NLL is defined as follows: $\sum_{i=1}^{N} -\log p\left(\hat{\mathrm{y}}_i | \mathrm{x}_{i}\right)$.


\noindent \textit{Brier Score:} The Brier score computes the $\ell_2$ metric between the predicted likelihoods and the true labels~\cite{degroot1983comparison,gneiting2007strictly}:
\begin{equation}
\label{eq:brier}
\mathrm{BS}= {\frac{1}{N}}{\sum_{n=1}^{N} \sum_{k=1}^{K}\left[p_\theta \left(\hat{\mathrm{y}}_{n}=k | x_{n}\right)-\mathbb{I}\left({\mathrm{y}}_{n}=k\right)\right]^{2}}
\end{equation}
\subsection{Impact of Calibration on Ticket Performance}
\subsubsection{(i) MNIST and Fashion-MNIST with a Fully Connected Network}
We conducted an initial investigation on the MNIST digit recognition~\cite{lecun2010mnist} and the Fashion-MNIST~\cite{xiao2017fashion} datasets using simple, fully connected networks (FCN). We adopt the architecture and hyper-parameters from~\cite{frankle2018lottery}, i.e., we use a LeNet-300-100 \cite{lecun1998gradient} as our base architecture for this experiment. The two layers in the network contained $300$ and $100$ neurons respectively. In the case of MNIST, we used a learning rate of $1e-3$ with the Adam optimizer \cite{ADAM} for $80$ epochs and using mini-batches of $60$. In the case of Fashion MNIST, we used mini-batches of size $128$ and trained for $90$ epochs.  Following~\cite{frankle2018lottery}, we adopted the local pruning strategy for both these datasets. In particular, we performed magnitude-based weight pruning to select the sparse sub-networks, and the pruning ratio was set to $20$\% in each iteration except for the last layer, which is pruned at $10$\%.



Figures~\ref{fig:perf}(a),~\ref{fig:perf}(b) show the results of different calibration strategies, in comparison to the standard LTH, on these two datasets. In particular, we report the accuracy and the three calibration metrics, averaged across three random trials. We find that all prediction calibration methods perform comparatively to the basic LTH; however with marginal improvements for tickets obtained with an explicit confidence calibration at lower pruning iterations. We surmise this is due to the low complexity of both the datasets and architecture considered. With simpler model architectures and classification tasks, it is highly likely the trained models are inherently well-calibrated and including an additional calibration objective does not lead to significant improvements. Interestingly, arbitrarily increasing the dropout rate for \textit{LWCC-SI} and \textit{VWCC} in this case led to a drop in the accuracies. However, the gains achieved by tickets from well-calibrated models on more complex models/data are non-trivial and can be evidenced from CIFAR-10 with ResNet-18 experiment.

\subsubsection{(iii) CIFAR-10 with ResNet-18}
In this experiment, we used the CIFAR-10~\cite{cifar10} with a ResNet-18~\cite{Resnet} model. Following~\cite{frankle2018lottery}, in this case, we performed global pruning at the ratio of $20$\% in each iteration, and we did not prune the parameters used for downsampling outputs from residual blocks or the final fully-connected layer.  
We trained the networks using the SGD optimizer at the learning rate of $0.01$, weight decay of $0.0001$ and a momentum of $0.9$, for $130$ epochs. We annealed the learning rate by $0.1$ after $80$ and $120$ epochs. 

Figure~\ref{fig:perf}(c) plot the performance of the lottery tickets obtained from models with difference calibration strategies. The first striking observation is that, unlike the MNIST/Fashion MNIST datasets, calibrated networks provide better performing sub-networks. With increased model complexity, we also observe consistent improvements in calibration at all compression ratios thus hinting that the structure of the sub-network plays a critical role in the generalization of tickets, in addition to the initialization strategy in LTH. 
We note that strategies that explicitly promote confidence calibration, namely \textit{VWCC} and  \textit{LWCC-SI}, and augmentation strategies such as \textit{Mixup} provide maximal benefits, while approaches that adjust the softmax probabilities with simplistic priors, e.g. uniform marginal distribution in \textit{MDA}, provide only marginal improvements.

\begin{figure*}[!t]
	\centering
	\subfloat[S][Brightness]{\includegraphics[width=0.48\linewidth,keepaspectratio]{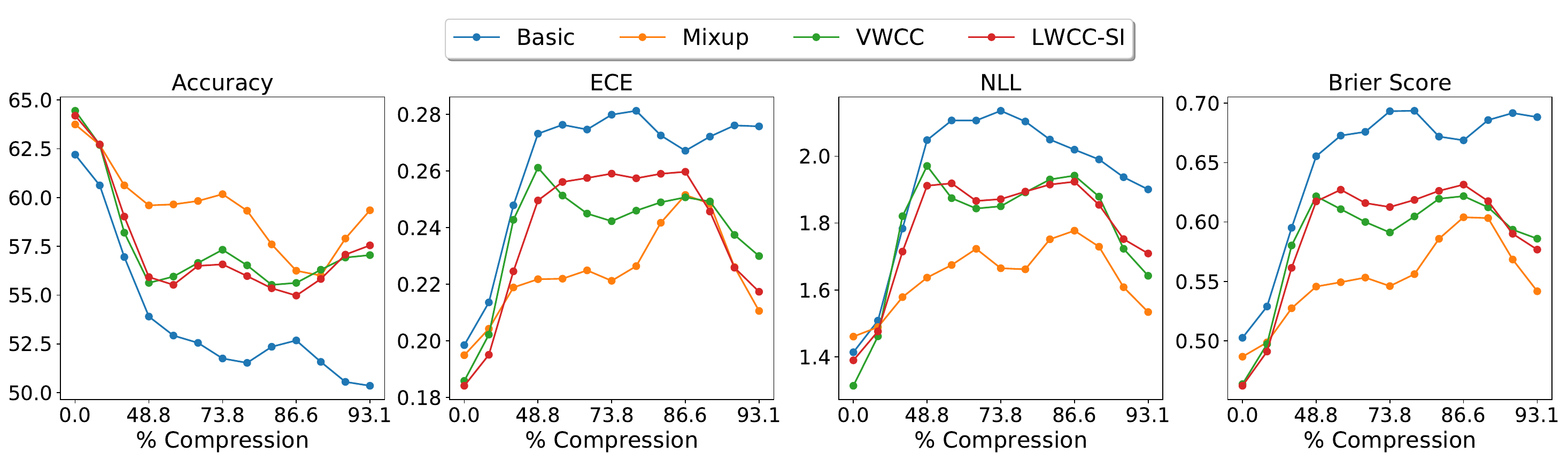}\label{fig:brightness}} \hfill
	\subfloat[S][Contrast]{\includegraphics[width=0.48\linewidth,keepaspectratio]{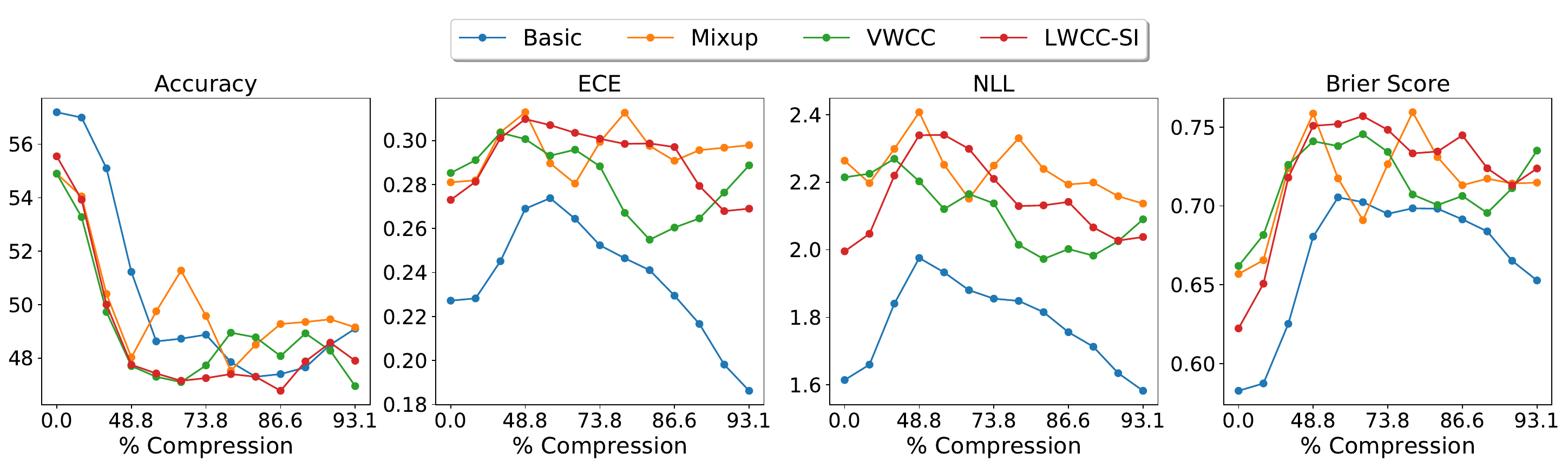} \label{fig:contrast}} \vfill
	\subfloat[S][Defocus Blur]{\includegraphics[width=0.48\linewidth,keepaspectratio]{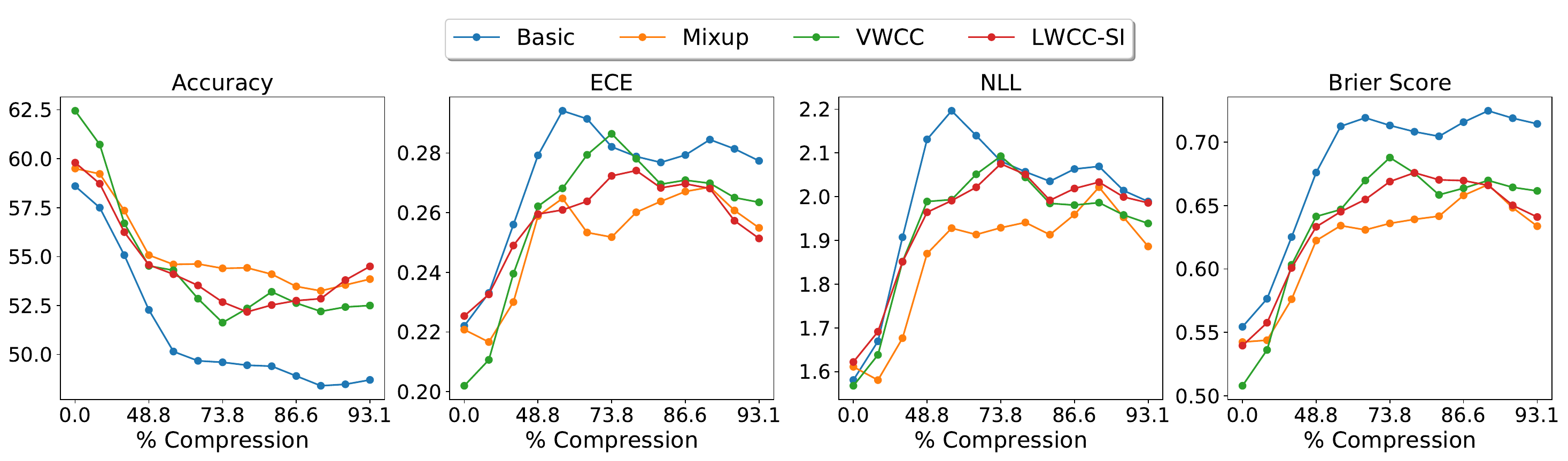} \label{fig:defocus}}\hfill
	\subfloat[S][Glass Blur]{\includegraphics[width=0.48\linewidth,keepaspectratio]{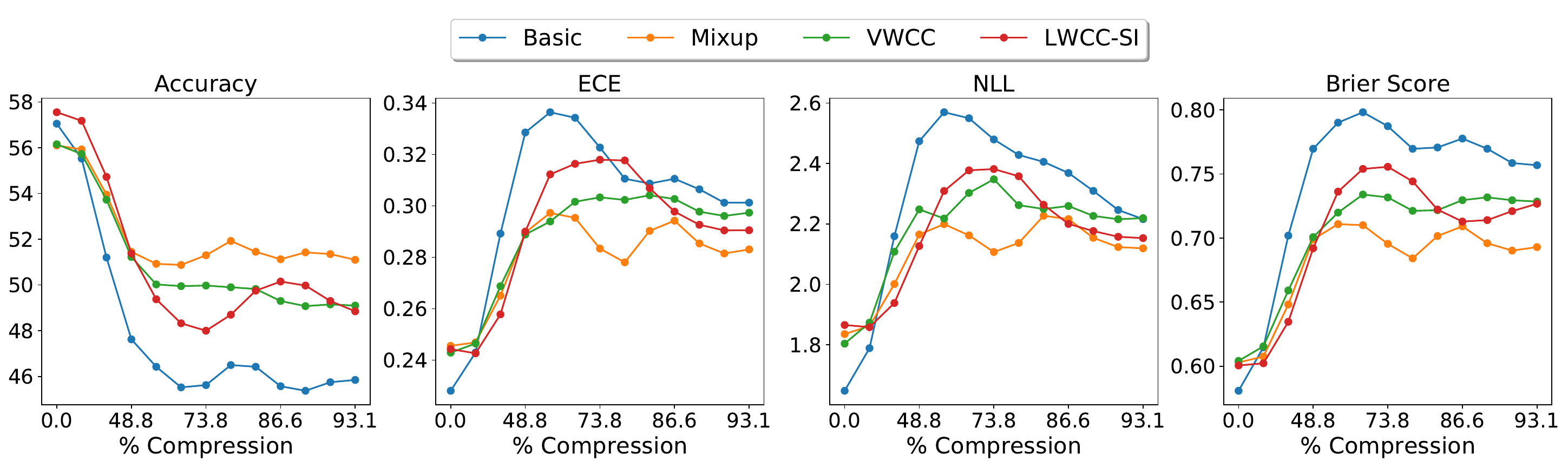} \label{fig:glass}}
	\vfill
	\subfloat[S][Fog]{\includegraphics[width=0.48\linewidth,keepaspectratio]{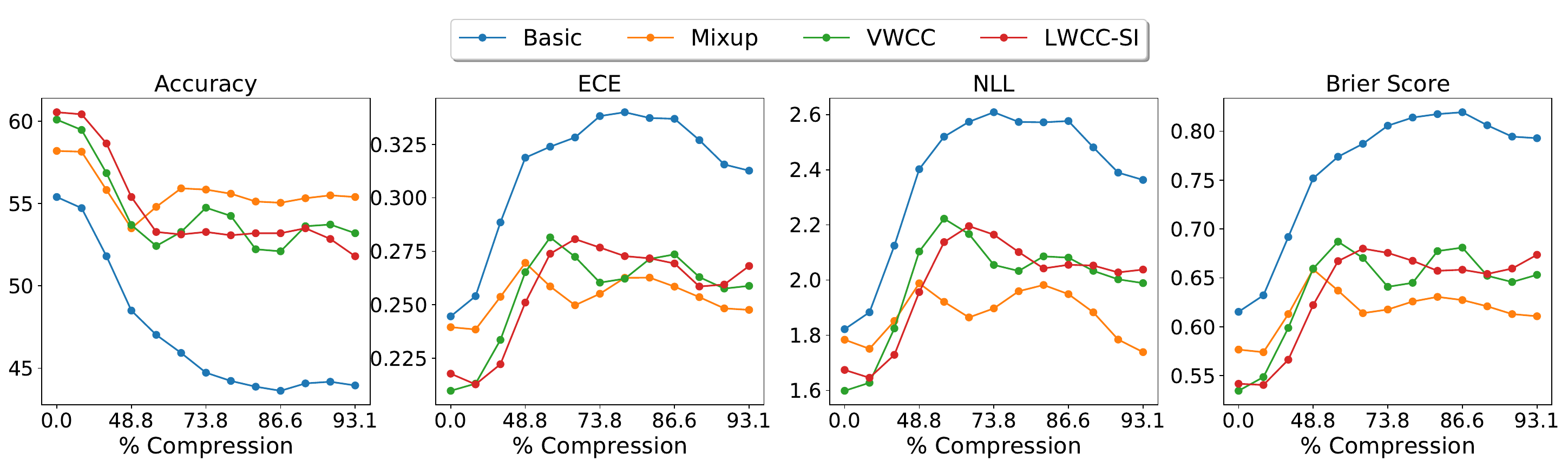} \label{fig:fog}}\hfill
	\subfloat[S][Frost]{\includegraphics[width=0.48\linewidth,keepaspectratio]{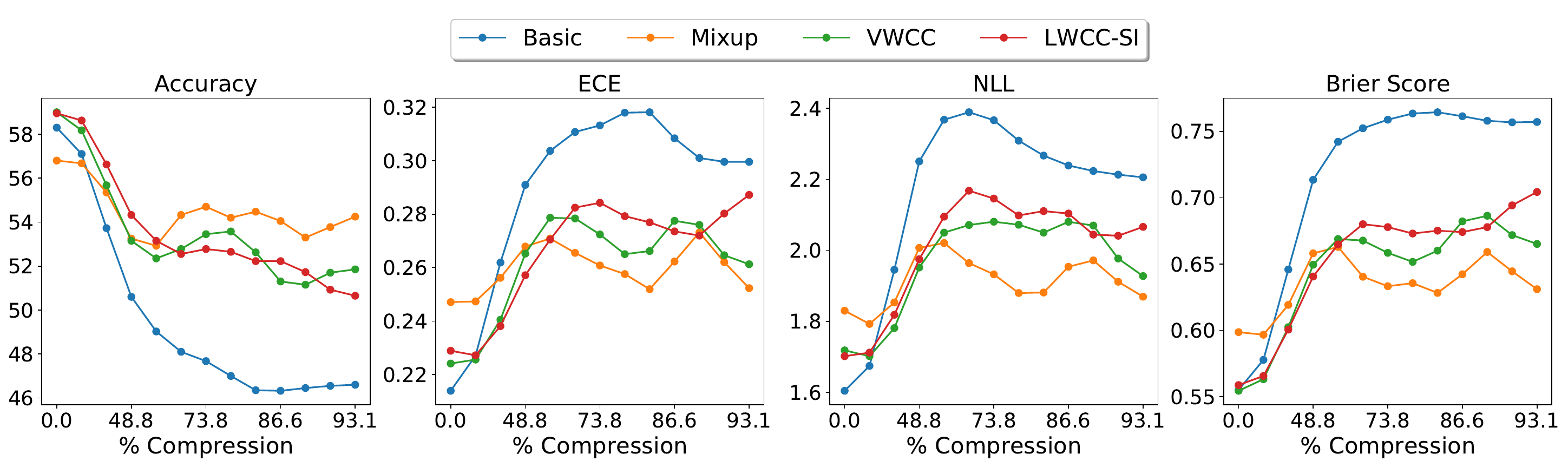} \label{fig:frost}}
	\vfill
	\subfloat[S][Snow]{\includegraphics[width=0.48\linewidth,keepaspectratio]{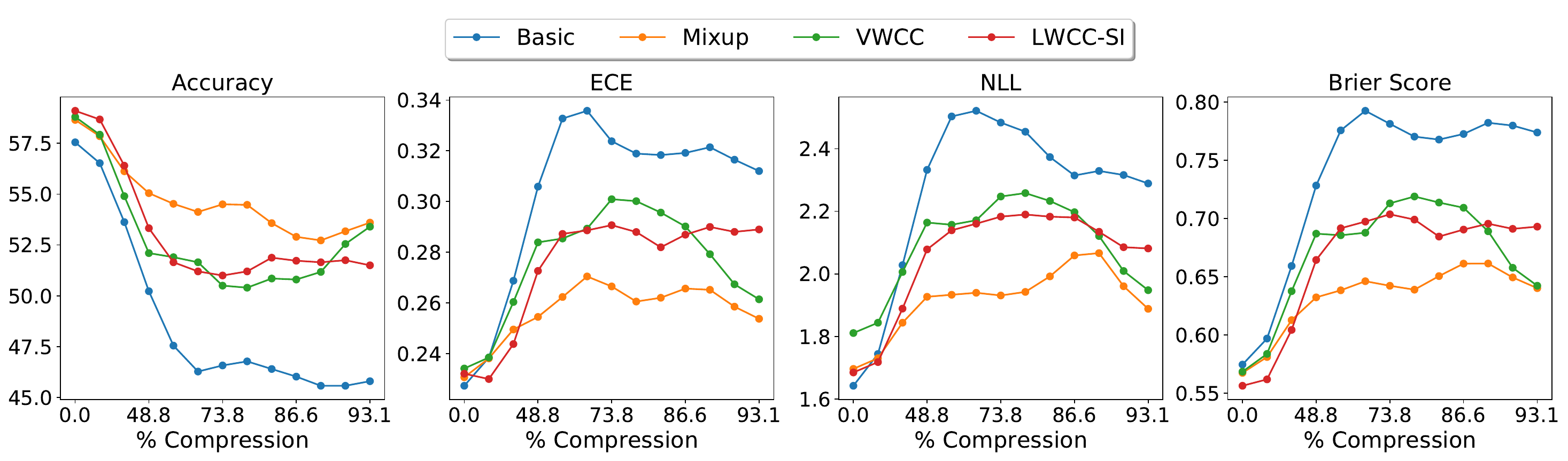} \label{fig:snow}}\hfill
	\subfloat[S][Motion Blur]{\includegraphics[width=0.48\linewidth,keepaspectratio]{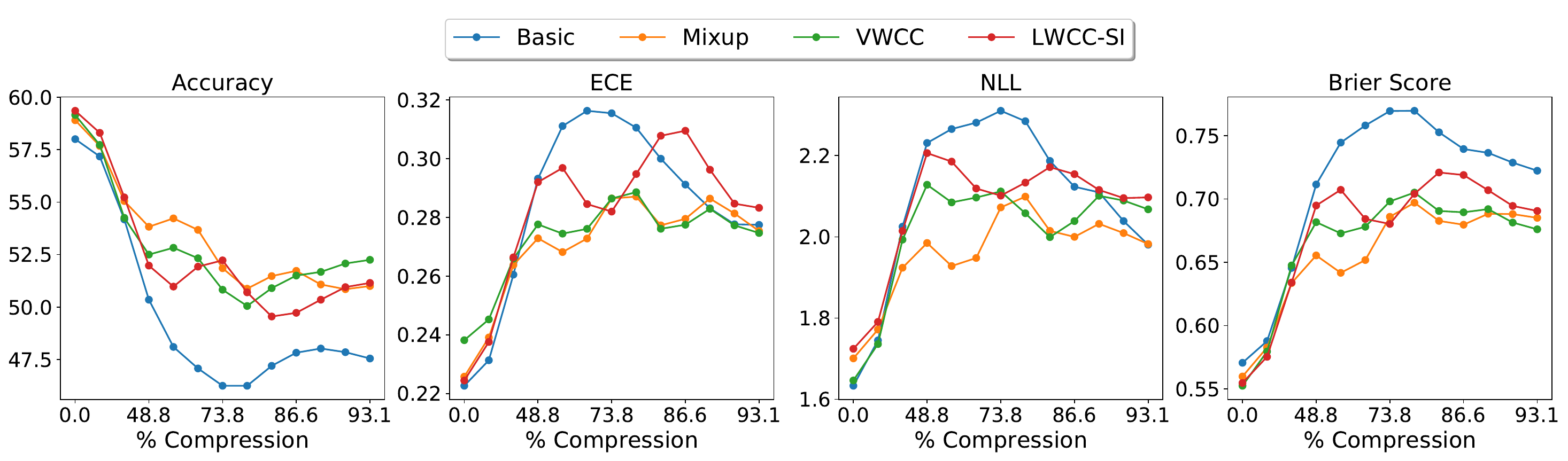} \label{fig:motion}}
	\vfill
	\subfloat[S][Gaussian Noise]{\includegraphics[width=0.48\linewidth,keepaspectratio]{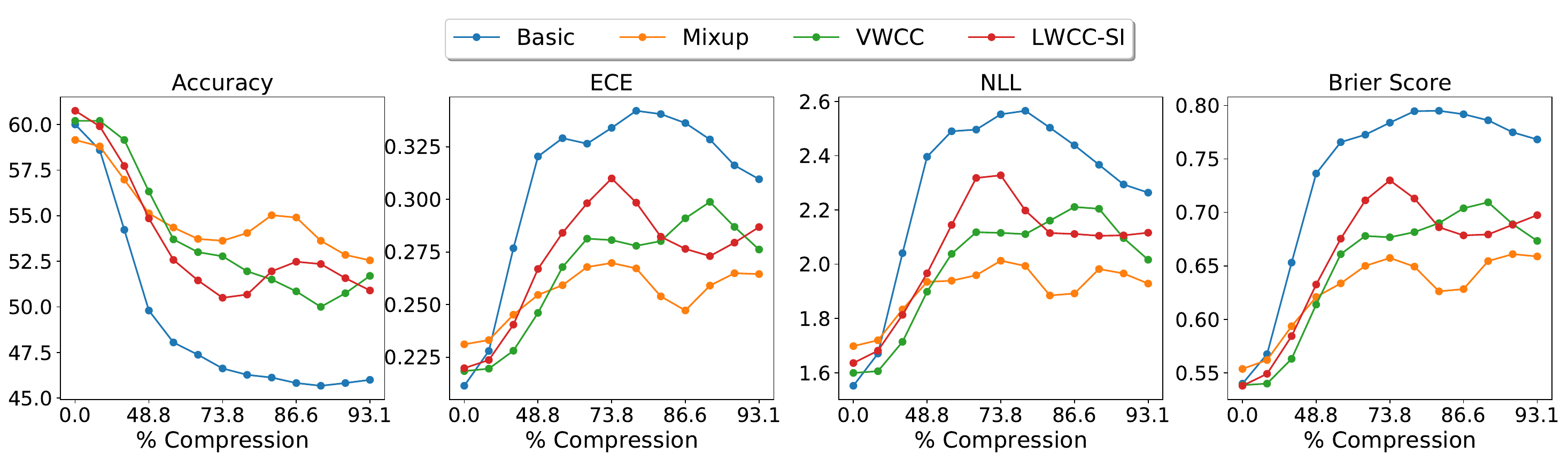} \label{fig:gaussian}}\hfill
	\subfloat[S][Shot Noise]{\includegraphics[width=0.48\linewidth,keepaspectratio]{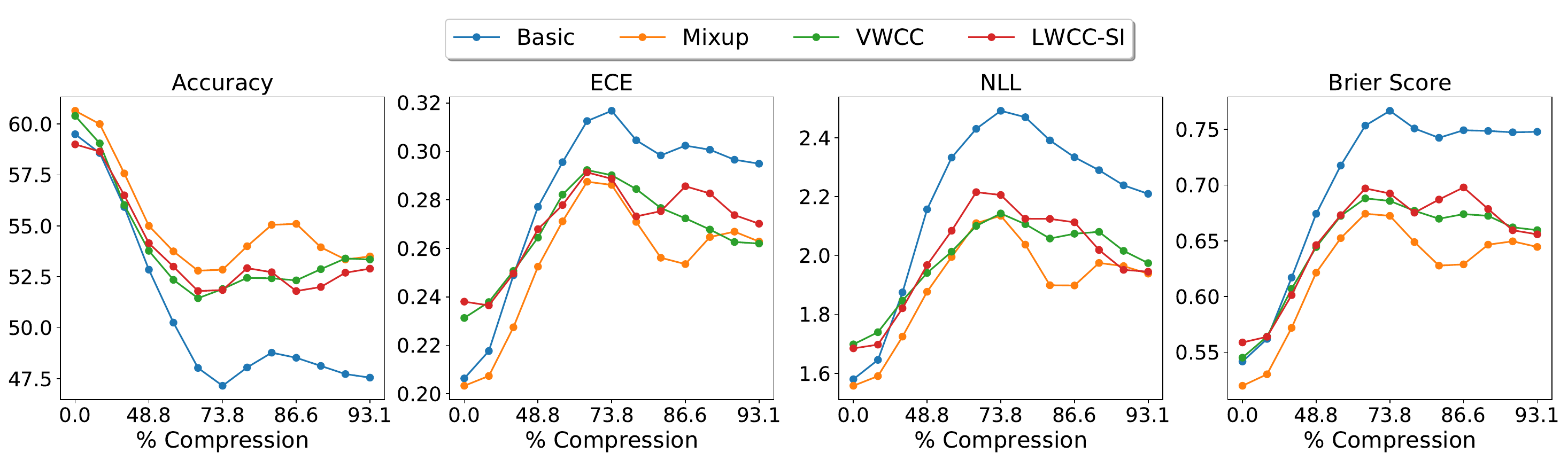} \label{fig:shot}}
	\vfill
	\subfloat[S][Pixelate]{\includegraphics[width=0.48\linewidth,keepaspectratio]{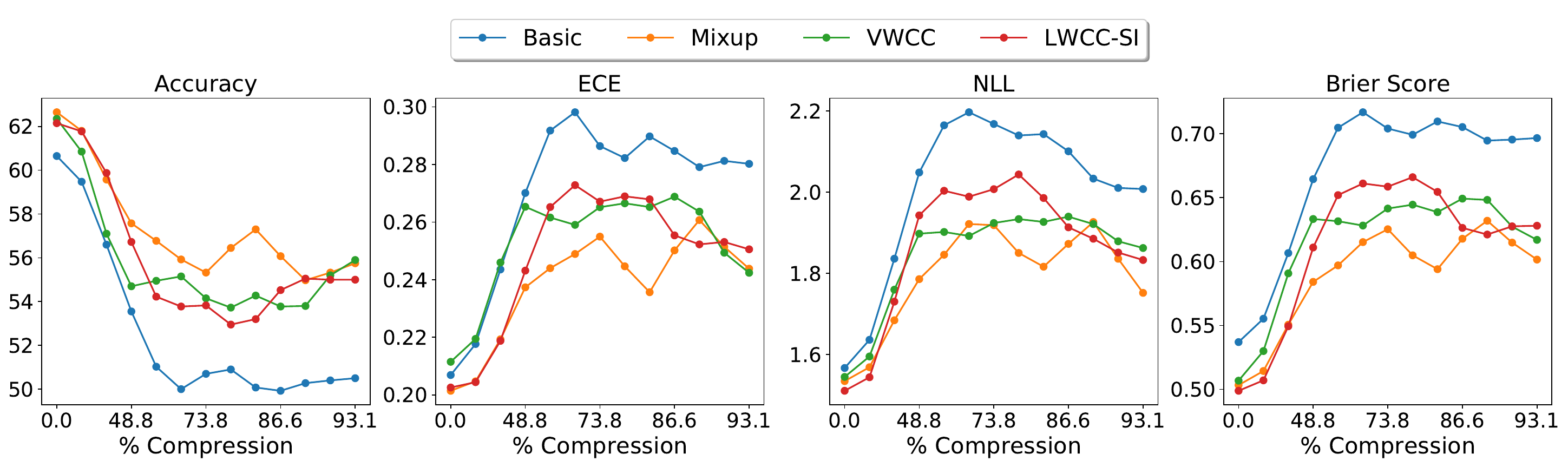} \label{fig:pixelate}}\hfill
	\subfloat[S][Jpeg Compression]{\includegraphics[width=0.48\linewidth,keepaspectratio]{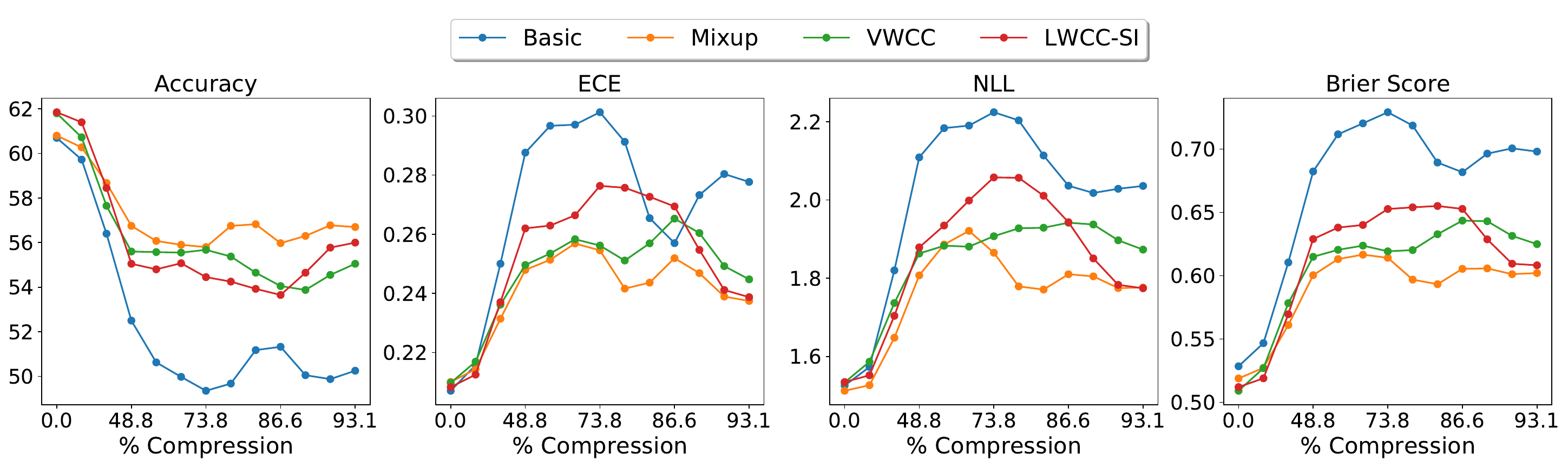} \label{fig:jpeg}}
	\caption{Ticket transfer performance on target datasets (CIFAR-10-C) that are characterized by distribution shifts when compared to the source data (standard CIFAR-10). The shifts were created using natural image corruptions, and we used ResNet-18 models for this experiment. We show the performance on the held-out test set for each of the corruptions.}
	\label{fig:cifar10c}
\end{figure*}

\subsection{Ticket Reusability under Distribution Shifts}
Prediction calibration in supervised learning is known to provide improved robustness under distribution shifts. In this section, we investigate if tickets from a source dataset are retrained using another target dataset, characterized by unknown shifts, will lead to improved performance than the standard LTH. Note that, we do not consider change in the task as assumed in the transfer learning experiments with LTH in previous works~\cite{oneticket}. Given the ability of confidence calibration to temper the influence of neurons that can potentially cause miscalibration, we expect our winning tickets to increasingly outperform LTH, as the degree of discrepancy between the source and target datasets increase. In order to test this hypothesis, we consider the two following experiments: (i) CIFAR-10a to CIFAR-10b benchmark~\cite{oneticket}, where the distribution shift caused only by sampling biases; (ii) CIFAR-10 to CIFAR-10C benchmarks, where the distribution shifts are caused by severe natural image corruptions. Similar to the empirical studies in the previous section, we evaluate the prediction performance and reliability of the resulting models through the three calibration metrics.



\subsubsection{(i) CIFAR-10a to CIFAR-10b}
Following the experimental setup in~\cite{oneticket}, we divide the CIFAR-10 dataset into two equal training splits namely CIFAR-10a and CIFAR-10b with $25$k training samples in each, with $2.5$k samples in each class. The source model was trained on the CIFAR-10a split and the CIFAR-10b set was treated as the target. Note that the distribution shift between the source and target datasets are solely due to sampling biases and is a relatively simpler shift to handle in practice. Following the CIFAR-10 experiment, we used the ResNet-18 architecture for both source and target models, and the hyperparameter settings for training both models were adopted from \cite{frankle2018lottery}. In this case, we used the SGD optimizer with learning rate 0.01, momentum 0.9 and weight decay 0.0001 and batch size 128. As mentioned earlier, we do not prune the fully connected layers and perform global pruning. Given the wining tickets from the source dataset, we retrain the model for the target dataset and evaluate the performance on the test set from CIFAR-10b. 

From Figure \ref{fig:cifar10ab}, we observe that the proposed approaches provide a bigger margin of improvement over basic LTH ($~1\%$) at all compression ratios, when compared to the $(~0.3$\% to $0.5\%)$ accuracy improvement in the case of CIFAR-10. This clearly indicates that, even with moderately severe distribution shift, the choice of the sub-network plays a very critical role in determining its effectiveness. In particular, we find that \textit{Mixup} and \textit{VWCC} calibration strategies provide the maximal gain.



\subsubsection{(ii) CIFAR-10 to CIFAR-10-C}
In this experiment, we retrain tickets from the clean CIFAR-10 source dataset to retrain on the challenging CIFAR-10C benchmark. Note that, the CIFAR-10-C dataset~\cite{hendrycks2019robustness} was created by applying 15 different natural image corruptions such as Gaussian noise, snow, fog, blur etc., to the CIFAR-10 test set. This dataset consists of $50$k samples, wherein each corruption is applied with five levels of severity. For our experiment, we considered a subset of $12$ corruptions including brightness, contrast, Gaussian noise, shot noise, glass blur etc.. We used the $10$k samples of CIFAR-10-C dataset, corresponding to level $5$ corruption, and created random train-test splits of $9$K and $1$K respectively. Following the CIFAR-10a to CIFAR-10b experiment, we used the same architecture and hyperparameter settings. 

For the sake of clarity, we only illustrate the best performing calibration methods, namely \textit{Mixup}, \textit{VWCC} and \textit{LWCC-SI}. As observed from Figure~\ref{fig:cifar10c}, the source winning tickets obtained from calibrated networks generalize significantly better in almost all cases, except under the \textit{Contrast} corruption. Interestingly, compared to CIFAR-10a to CIFAR-10b experiment, the distribution shifts here are significantly more severe and confidence calibration leads to orders of magnitude improvements in the performance. For example, in the cases of fog, frost or snow corruptions, we observe even $10\%-12\%$ improvements over the standard LTH tickets (even at higher compression ratios). In addition to analyzing the accuracies of the target models trained using the source tickets, we evaluated the reliability of the resulting models. Similar to our previous empirical studies, we find that our approach leads to much improved calibration scores in all cases. These results clearly evidence the importance of including confidence calibration into the model training process, particularly when retrained under challenging distribution shifts. In the next section, we summarize all our key findings and provide recommendations for improving lottery tickets in practice.


\section{Key Findings}

\begin{itemize}
\item While different pruning strategies have been explored in existing works~\cite{zhou2019deconstructing}, the common conclusion has been that weight magnitude based pruning is the most effective, and hence the research focus has shifted towards investigating better initialization strategies for the sub-networks. However, our results clearly show that using prediction calibration during the training of the over-parameterized model can produce sub-networks that demonstrate improved generalization (under distribution shifts) and produce consistently reliable models (showed using calibration metric evaluations on different dataset/model combinations). This is an interesting result in that we have resorted to the vanilla initialization strategy adopted by LTH~\cite{frankle2018lottery} and the performance improvements are solely from more effective sub-networks. This motivates further research to better understand the role of the sub-network selection, not by merely adjusting the pruning criterion, but by designing networks that are not just accurate but also better calibrated to meaningful priors.  

\item In cases of simpler classification tasks such as Fashion-MNIST or MNIST, we find that using confidence calibration provided only minor improvements over tickets from models with no explicit calibration. Interestingly, we also noticed that, strengthening the regularization (e.g., increasing the dropout rate in \textit{VWCC}) on already well-calibrated models led to inferior performance, implying effects over-regularization. In contrast, with challenging tasks such as CIFAR-10 classification, prediction calibration consistently led to improved tickets. 

\item The most important observation is that even under challenging distribution shifts, i.e. CIFAR-10 to CIFAR-10-C experiment, the tickets obtained from models with an explicit calibration objective showed consistently superior performance when compared to the source tickets obtained using standard LTH, clearly evidencing the vulnerabilities of miscalibrated models and tickets inferred from them. 

\item Our results are particularly important in the context of recent efforts that attempt to design randomly initialized neural networks that can be utilized for a given dataset, without even carrying out model training~\cite{ramanujan2019s,gaier2019weight}. While sufficiently over-parameterized random networks will most likely contain sub-networks that achieve reasonable accuracy without training, calibration strategies can help identify the most effective, in terms of both generalization and reliability.

\end{itemize}

 \section{Acknowledgements}
This work was performed under the auspices of the U.S. Department of Energy by Lawrence Livermore National Laboratory under Contract DE-AC52-07NA27344. 


\begin{quote}
\begin{small}
\bibliography{references}
\bibliographystyle{aaai21}
\end{small}
\end{quote}

\end{document}